\documentclass[10pt,twocolumn,letterpaper]{article}

\usepackage{wacv}
\usepackage{times}
\usepackage{epsfig}
\usepackage{graphicx}
\usepackage{amsmath}
\usepackage{amssymb}
\usepackage{booktabs}

\usepackage{stfloats}
\usepackage{booktabs}
\usepackage[lined, boxed, ruled]{algorithm2e}

\def\wacvPaperID{****} % Enter the WACV Paper ID here

\wacvalgorithmstrack   % Uncomment this line if you are submitting to the Algorithms Track.
\wacvfinalcopy 
\ifwacvfinal
\usepackage[breaklinks=true,bookmarks=false]{hyperref}
\else
\usepackage[pagebackref=true,breaklinks=true,colorlinks,bookmarks=false]{hyperref}
\fi

\begin{document}

%%%%%%%%% TITLE

\title{Toward a Deeper Understanding: RetNet Viewed through Convolution}

\author{Chenghao Li\\
KAIST, South Korea\\
{\tt\small lch17692405449@gmail.com}
% For a paper whose authors are all at the same institution,
% omit the following lines up until the closing ``}''.
% Additional authors and addresses can be added with ``\and'',
% just like the second author.
% To save space, use either the email address or home page, not both
\and
Chaoning Zhang\thanks{Corresponding author}\\
Kyung Hee University, South Korea\\
{\tt\small chaoningzhang1990@gmail.com}
}

\maketitle
\thispagestyle{empty}

%%%%%%%%% ABSTRACT
\begin{abstract}
    The success of Vision Transformer (ViT) has been widely reported on a wide range of image recognition tasks. ViT can learn global dependencies superior to CNN, yet CNN's inherent locality can substitute for expensive training resources. Recently, the outstanding performance of  RetNet in the field of language modeling has garnered attention, surpassing that of the Transformer with explicit local modeling, shifting researchers' focus towards Transformers in the CV field. This paper investigates the effectiveness of RetNet from a CNN perspective and presents a variant of RetNet tailored to the visual domain. Similar to RetNet we improves ViT's local modeling by applying a weight mask on the original self-attention matrix. A straightforward way to locally adapt the self-attention matrix can be realized by an element-wise learnable weight mask (ELM), for which our preliminary results show promising results. However, the element-wise simple learnable weight mask not only induces a non-trivial additional parameter overhead but also increases the optimization complexity. To this end, this work proposes a novel Gaussian mixture mask (GMM) in which one mask only has two learnable parameters and it can be conveniently used in any ViT variants whose attention mechanism allows the use of masks. Experimental results on multiple small datasets demonstrate that the effectiveness of our proposed Gaussian mask for boosting ViTs for free (almost zero additional parameter or computation cost). Our code can be publicly available at \href{https://github.com/CatworldLee/Gaussian-Mixture-Mask-Attention}{https://github.com/CatworldLee/Gaussian-Mixture-Mask-Attention}.
\end{abstract}

%%%%%%%%% BODY TEXT

\section{Introduction}

\begin{figure}[t]
\centering
\includegraphics[width=\linewidth]{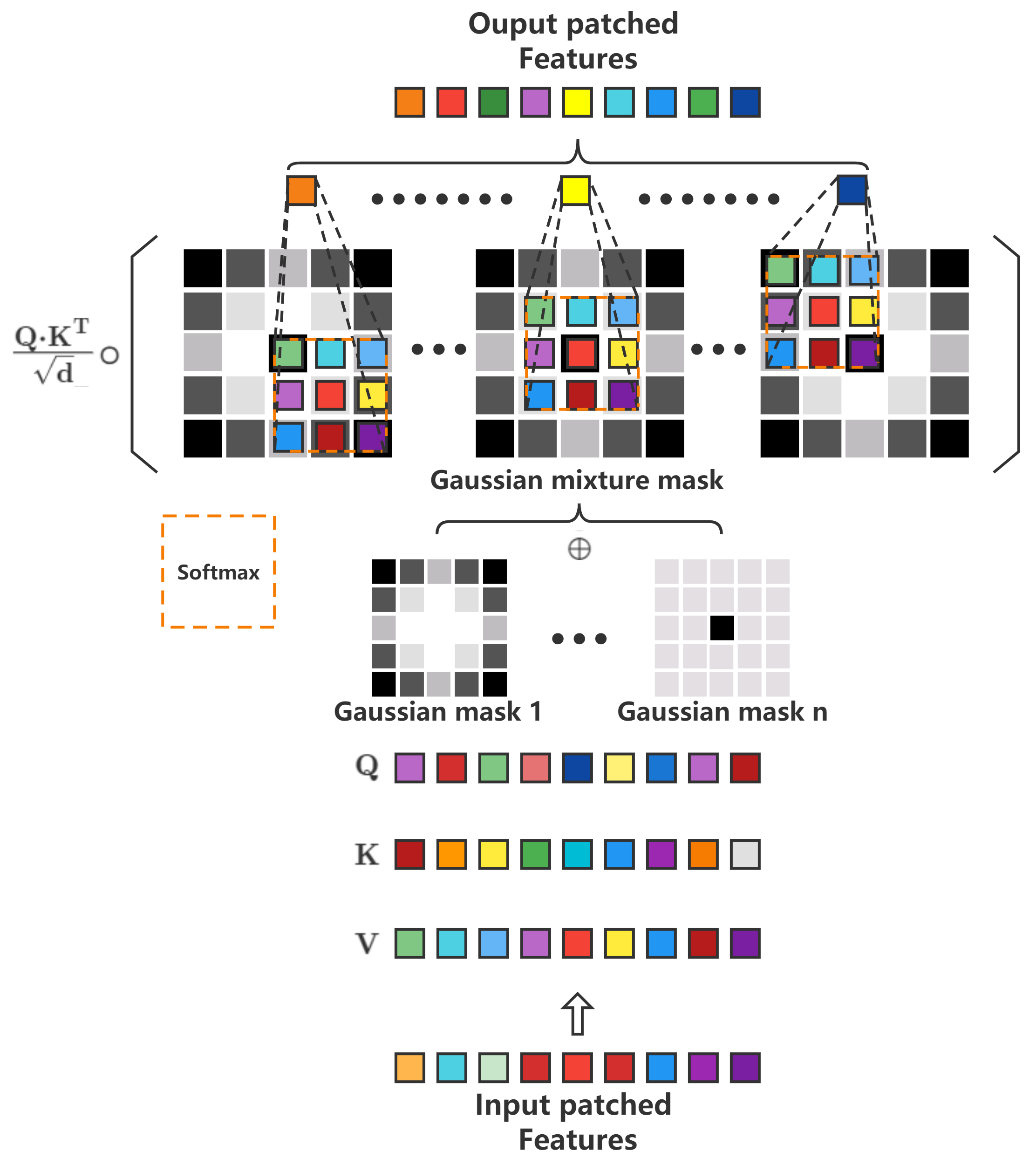}
\caption{\textbf{Overview of Gaussian Mixture Mask (GMM) Attention Mechanism.} Firstly, feature vectors are mapped into three matrices $Q$, $K$, and $V$ within the attention module (\textbf{bottom}). Subsequently, n distinct Gaussian masks are defined, which are linearly combined to form a Gaussian mixture mask. Building upon the foundational self-attention mechanism (\textbf{middle}), the shift window of the Gaussian mixture mask is unfolded and expanded into corresponding attention scores, this step resembles the feature vector undergoing an \textbf{element-wise convolution} operation, resulting in the attention map for each patch. Finally, the output patch feature is calculated as the dot product of the matrix $V$ and the attention map (\textbf{top}).}
\label{fig:img1}
\end{figure}

Since the success of AlexNet~\cite{krizhevsky2017imagenet}, Convolutional Neural Networks (CNNs) have become the standard for computer vision. Krizhevsky~\textit{et al.} show that convolutions are advantageous in visual tasks due to their invariance to spatial translations and their low correlative-inductive bias. Convolutions leverage three important concepts to achieve their effects: \textit{sparse interactions}, \textit{weight sharing}, and \textit{equivariant representations}~\cite{goodfellow2016deep}. On the other hand, transformers are becoming increasingly popular and are a focus of modern machine learning research. Since \emph{``Attention is All You Need''}~\cite{vaswani2017attention}, the research community has noticed an upsurge in transformer- and attention-based research~\cite{devlin2018bert, radford2018improving, chen2020generative}. Transformers are designed specifically for sequence modeling and translation tasks, and their significant feature is the use of attention models to model long-distance dependencies in the data. Its huge success in the language domain has prompted researchers to explore transformers in the computer vision domain~\cite{dosovitskiy2020image, cordonnier2019relationship, liu2021swin, wang2022pvt, gong2021vision, he2022masked}. With the competitive modeling capacity, visual transformers have achieved remarkable performance improvements compared to CNNs on multiple benchmark tests.

Recently, RetNet~\cite{sun2023retentive} has emerged as a potent successor to Transformer in the realm of large-scale language models by embracing three computational paradigms: \textit{parallel}, \textit{recurrent}, and \textit{chunkwise recurrent}. Concurrently, Fan~\textit{et al.}~\cite{fan2023rmt} attempts have been made in the literature to introduce the RetNet paradigm into the visual domain. This study endeavors to understand RetNet from a novel perspective of CNN and presents a new variant of the ViT paradigm that integrates the essence of RetNet.

The intrinsic local inductive bias inherent in CNNs is not possessed by Transformers. CNNs have gained an advantage on spatially strong natural dataset, such as small-scale imagery and video, by employing their formidable inherent local modeling capability~\cite{liu2021efficient}. While Transformers can learn this inductive bias from scratch on large dataset, such as ImageNet~\cite{dosovitskiy2020image} with little effort, it usually cannot attain the performance of CNNs on small-scale dataset. A common technique to tackle the small-data problem is to utilize the paradigm of pre-training followed by fine-tuning~\cite{takashima2023visual}, enabling the model to incorporate inductive bias as well as data distribution adaptation. Nevertheless, distributions for some small datasets departs far from the mainstream situation, as with medical image datasets~\cite{valanarasu2021medical}, highlighting the importance of training ViT from scratch on small datasets. It has been demonstrated that, by explicitly modeling locality in ViT can improve its performance on smaller datasets~\cite{liu2021efficient,lee2021vision, hassani2021escaping}.

Attention mechanism allows the models to automatically learn associations from the input. The Attention mechanism establishes a weight matrix, i.e. the Attention score introduced by the sofmax operation, to automatically adjust the affinity information between patches. RetNet introduces a pivotal change by replacing the softmax operation, essential for Transformers' self-attention, with a Hadamard product and an innovative D-matrix, followed by GroupNorm~\cite{sun2023retentive}.
In order to enhance the sharpness of the attention scores distribution, Lee~\textit{et al.}~\cite{lee2021vision} propose adding a learnable temperature parameter after the attention scores.  We propose an even more direct approach to obtain better results by adding an element-wise learnable mask (ELM) after the attention scores. The results demonstrate that the resulting model performs remarkably well. As a cost of improving the local modeling capability of a model, such element-wise learnable masks will greatly increase the number of parameters to be trained, resulting in more resource consumption of computing power and time.

In our preliminary experiment, element-wise learnable masks showed two characteristics: \textit{locality} and \textit{extroversion} in their training results, subsequently revealing two major issues of the self-attention mechanisms of ViTs: 

\begin{itemize}
    \item \textit{Locality}: The dependencies between adjacent patches become more obvious. Insensitive to image spatial information. Images have a stronger locality than text. This locality changes with the depth of the network, showing a stronger local correlation in shallow networks and stronger global correlation in deep networks~\cite{dosovitskiy2020image,li2022locality}. Although the position encoding in ViT can meet some spatial information requirements, it cannot meet more complex spatial information requirements. Attention Score in the attention mechanism requires stronger spatial information modeling ability.
    \item \textit{Extroversion}: Masks tend to suppress the impact of patches upon themselves, prompting the patches to be more reliant on other patches in the attention layers thus accelerating the flow of information. In the original ViT, the flow of information between layers is rather sluggish, as residual links preclude the model from transferring its information across patches which leads to a decrease in speed in the process of information iteration. Moreover, as the feature vector dimension is increased, the gradient of the softmax function brought on by the self-attention mechanism is disproportionately small, slowing down the learning process and further deepening the aforementioned speed problem. The addition of a scaling factor for the self-attention mechanism~\cite{vaswani2017attention} alleviates this issue to some extent, yet there is still room to improve.
\end{itemize}

To this end, we propose a Gaussian mixture mask (GMM), a dynamic learnable mask which implicitly generates a mask by learning two parameters $\sigma$ and $\alpha$ to modulate the locality of the attention mechanism. Research has confirmed that a particular case of Gaussian masking can be used to represent extroversion. Compared to the ELM, GMM utilizes very few parameters, and achieves better performance. In addition, theoretically speaking, as it is plug-and-play, GMM can be applicable to any variants of ViT leveraging self-attention, such as Swin~\cite{liu2021swin} and CaiT~\cite{touvron2021going}. In summary, our contributions are as follows: 
\begin{enumerate}
    \item We propose an Element-wise Learnable Mask  to identify two features of Vision Transformer's self-attention mechanism on small datasets.
    \item We propose a Gaussian Mixture Mask to boost ViTs for free on small datasets.
\end{enumerate}

\section{Related Works}

\subsection{Visual Transformers}
Recently, with the remarkable development of transformers in the field of NLP, many works have sought to introduce visual transformers into image classification. Cordonnier~\textit{et al.}~\cite{cordonnier2019relationship} created an original visual transformer composed of a standard transformer and a secondary position encoding. Subsequently, Vision Transformer~\cite{dosovitskiy2020image} introduced the transformer model and achieved excellent performance on the large dataset Imagenet. Following this, variants of hybrid prior knowledge emerged, including the Swin Transformer~\cite{liu2021swin}, which adopted a sliding window approach to achieve local and global modeling to gain multi-scale information while reducing computational complexity. To address the feature granularity neglected in ViT and the high cost of computation, researchers developed Hierarchical Transformers~\cite{yuan2021tokens,wang2021pyramid,heo2021rethinking,wang2022pvt,wu2021cvt}, which adopted a hierarchical modeling approach: T2T utilized the overlapping expansion operations, while PiT and CvT utilized pooling and convolution to down-sample. Subsequently, Deep Transformers~\cite{touvron2021going,zhou2021deepvit,gong2021vision,2106.03714} were devoted to improving the learning capability by increasing the model depth, and Transformers with Self-Supervised Learning~\cite{chen2020generative,li2021mst,bao2021beit,he2022masked,chen2021empirical,caron2021emerging,xie2021self} transferred unsupervised training to the visual transformer field.

\subsection{Localness Modeling on Transformers}

Standard Transformer models already possess implicit or explicit local modeling which is accomplished through learnable or non-learnable position embeddings, leading Transformer to be more inclined towards local patches. Recent research in NLP fields indicates that explicit local modeling can further improve model performance~\cite{sun2023retentive,shaw2018self,kim2020t,fan2021mask}. RetNet~\cite{sun2023retentive} employs explicit distance decay for local modeling and has emerged as a formidable successor to Transformer in large-scale language models. ASR~\cite{sperber2018self} has showed that localized long sequence representations perform better in both speech modeling and natural language inference tasks in self-attention models.Moreover, T-GSA~\cite{kim2020t}'s Gaussian distance model offers superior performance compared to the prevailing Transformer-based recurrent models and LSTMs. DMAN~\cite{fan2021mask} proposed a dynamic masking attention network with a learnable mask matrix that performs local modeling in an adaptive manner. In the CV field, there are also researches improved by explicit local modeling~\cite{fan2023rmt,cheng2022masked}. RMT~\cite{fan2023rmt} introduces RetNet into the visual domain and combines it with the Vision Transformer, offering a model mechanism, Retentive Self-Attention (ReSA), endowed with spatial locality prior knowledge. Mask2Former~\cite{cheng2022masked} takes advantage of masks to locally restrict attention, and thereby reduces the research workload while improving model performance on image segmentation tasks. Notably, locality is also used to balance the modeling capability and the efficiency of Transformer~\cite{beltagy2020longformer,zaheer2020big}. Longformer~\cite{beltagy2020longformer} is proposed to address the limitation of Transformer models on long sequence processing. Its attention mechanism can be linearly related to the sequence length, and thus readily tackle documents of thousands of tokens or longer. BIGBIRD~\cite{zaheer2020big} leverages sparse attention mechanisms, including local attention, which is inspired by the sparse techniques from the graph structure, and reduces the complexity down to linear, i.e., O(N).

\subsection{Visual Transformers on Small Datasets}

Recently, several methods ~\cite{liu2021efficient, hassani2021escaping, touvron2021training, lee2021vision, li2022locality} have been explored to enhance ViT on small datasets. Liu~\textit{et al.}~\cite{liu2021efficient} proposed an auxiliary self-supervision task to extract extra information from images, thereby effectively improving the training of ViT with few samples. Hassani~\textit{et al.}~\cite{hassani2021escaping} proposed a new model architecture to boost the performance of ViT on small datasets, including utilizing small patch size, introducing convolution in shallow layers, and dropping classification tokens. Touvron~\textit{et al.}~\cite{touvron2021going} proposed a distillation method called DeiT, whose core idea is to upscale using a convolutional network as a teacher network, which presents better performance compared to using a Transformer-structured network as a teacher network. Li~\textit{et al.}~\cite{li2022locality} proposed a similar method to DeiT, introducing local knowledge distillation to improve the performance of ViTs model on small datasets, which achieves local guiding through imitating a training convolutional neural network, making ViTs model converge faster and significantly improving the performance on small datasets. The work most closely resembling ours is that of Lee~\textit{et al.}~\cite{lee2021vision}, who proposed Shift Patch Tokenization (SPT) and Locality Self-Attention (LSA) to effectively address the absence of local inductive bias in ViT. SPT embeds the spatial information between adjacent pixels to provide a wider receptive field to visual tokens. LSA remedies or alleviates the flatness problem of the attention scores by adding a learnable temperature and diagonal masking. In this work, we proposed an Element-wise simple Learnable Mask (ELM) as a generalization version of the LSA introduced by Lee~\textit{et al.} which proves to perform better. Furthermore, we also introduced a Gaussian Mixture Mask (GMM) for further performance improvement with reduced parameters and computation amid no extra cost.

\section{Approach}\label{approach}

In this part, we first revisit the attention mask from the perspective of convolution. Subsequently, we introduce an Element-wise Mask (ELM) and analyze two properties of it in Section~\ref{sec: SLM}, followed by an elaboration of the definition of Gaussian Mixture Mask (GMM) in Section~\ref{sec:GMM}, together with an explanation on how GMM modifies the attention mechanism in ViTs in Section~\ref{sec:GMMAtten}, accompanied by the pseudo-code of the algorithm.\\

\begin{figure*}[t] 
\centering 
\includegraphics[width=\textwidth]{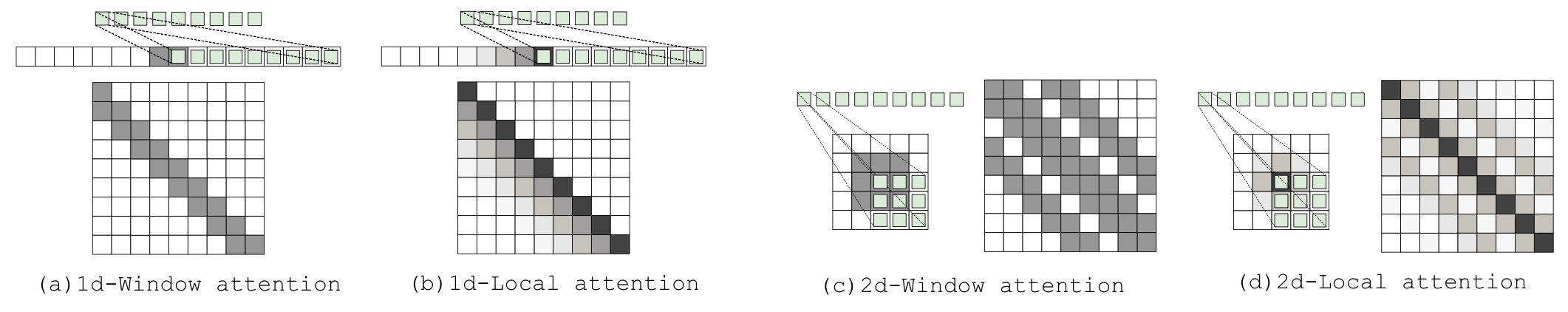} 
\caption{Four examples are provided to intuitively demonstrate this convolution operation. (a) Windowed attention in the 1d case. (b) Local attention in the 1d case. (c) Windowed attention in the 2d case. (d) Local attention in the 2d case.} 
\label{fig:conv}
\end{figure*}

\noindent\textbf{Element-wise Conv Operation.} If the mask is a circulant matrix with multi-diagonal properties, when such a matrix is added to the attention scores using the Hadamard product, it can be perceived as performing an operation akin to an element-wise convolution. The \textit{convolution kernel} is the hidden feature vector, while the \textit{feature map} is the weight mask. Examples of 1d and 2d cases are shown in the figure~\ref{fig:conv}.

\subsection{Element-wise Learnable Mask}\label{sec: SLM}

In our preliminary investigation, an element-wise learnable matrix which has the same shape as the attention score matrix $\left(N_{patches} \times N_{patches}\right)$ was added behind the attention score matrix. The value of its $i$-th row and $j$-th column indicates the preference of the feature information in the $i$-th patch of the subsequent layer for the information of the $j$-th patch in the current layer. This mask layer is aimed at dynamically adjusting the tendency of patch attention. After performing numerous experiments, we discovered that the element-wise learnable mask mainly exhibits two tendencies: \textit{locality} and \textit{extroversion}.

\begin{enumerate}

    \item \textit{Locality}: Element-wise learnable masks are always characterized by a high degree of locality. In shallow networks of ViT, the tendencies of masks are localized, and masks encourage patches to absorb information from neighboring patches. In deep networks, masks are tend to be more globalized, and masks encourage patches to absorb information from far away patches. Additionally, we find that masks not only promote the absorption of information from adjacent patches, but could show a counter-tendency, i.e., masks inhibit the absorption of adjacent information and absorb information from a distance in some cases. However, regardless of their promotion or inhibition, masks always vary according to the distance between patches.

    \item \textit{Extroversion}: Element-wise learnable masks frequently display suppressive patches that depend on their own information.  Whether the mask strongly promotes or suppresses patches to depend on the surrounding patches, it always suppresses patches to rely on themselves.  This suppressive phenomenon does not vary with the distance between patches and the amount of patches.   This extroversion can be considered as frequently appearing locality cases that have extremely close distances.
\end{enumerate}

\begin{figure}[h]
    \centering
    \includegraphics[width=\linewidth]{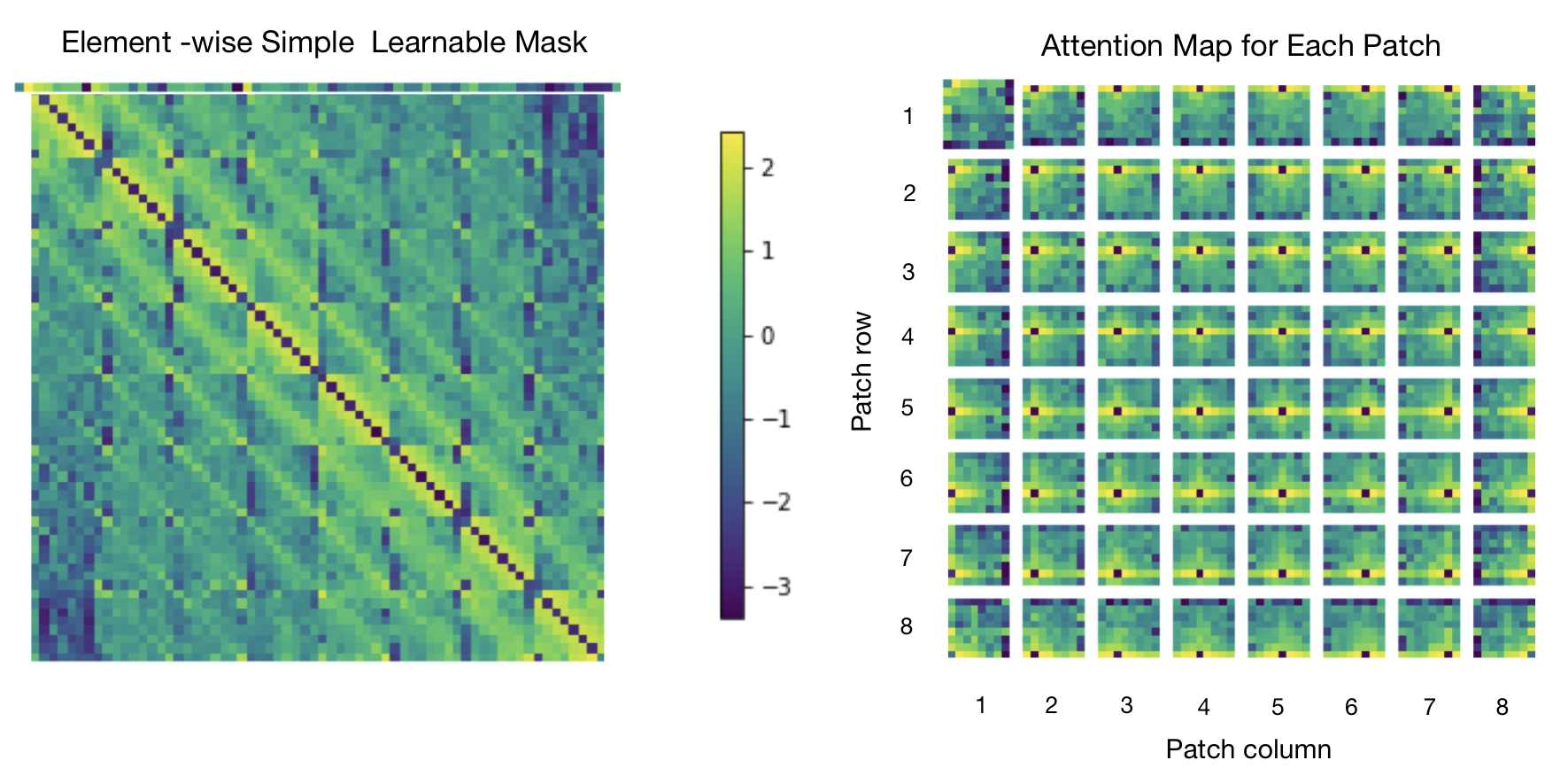}
    \caption{\textbf{Left:} 
    A 64$\times$64 simple yet learnable element-wise mask trained from a Tiny-ViT on CIFAR-10 dataset reaching up to 94.02\% Top-1 accuracy, is presented. Each pixel in the mask represents the value in the mask with brighter areas having higher values and darker areas having lower values. \textbf{Right:} The left figure is folded along the row into 64 attention maps, and then arranged into an $8 \times 8$ grid, with each grid consisting of an $8 \times 8$ pixel attention map.  In the right figure, the attention map in each grid represents the attention map of the corresponding patch position.}
    \label{fig:attention_map}
\end{figure}

The illustration in Fig.~\ref{fig:attention_map} presents above two patterns, the single-layer ELM on the left, and an attention map of each patch position on the right. Observably, the concentration of the brightness surrounding each patch decays with distance, with the patch itself appearing particularly faint. This reflects the phenomenon of both locality and extroversion.

\begin{figure*}[t]
    \centering
    \includegraphics[width=0.9\linewidth]{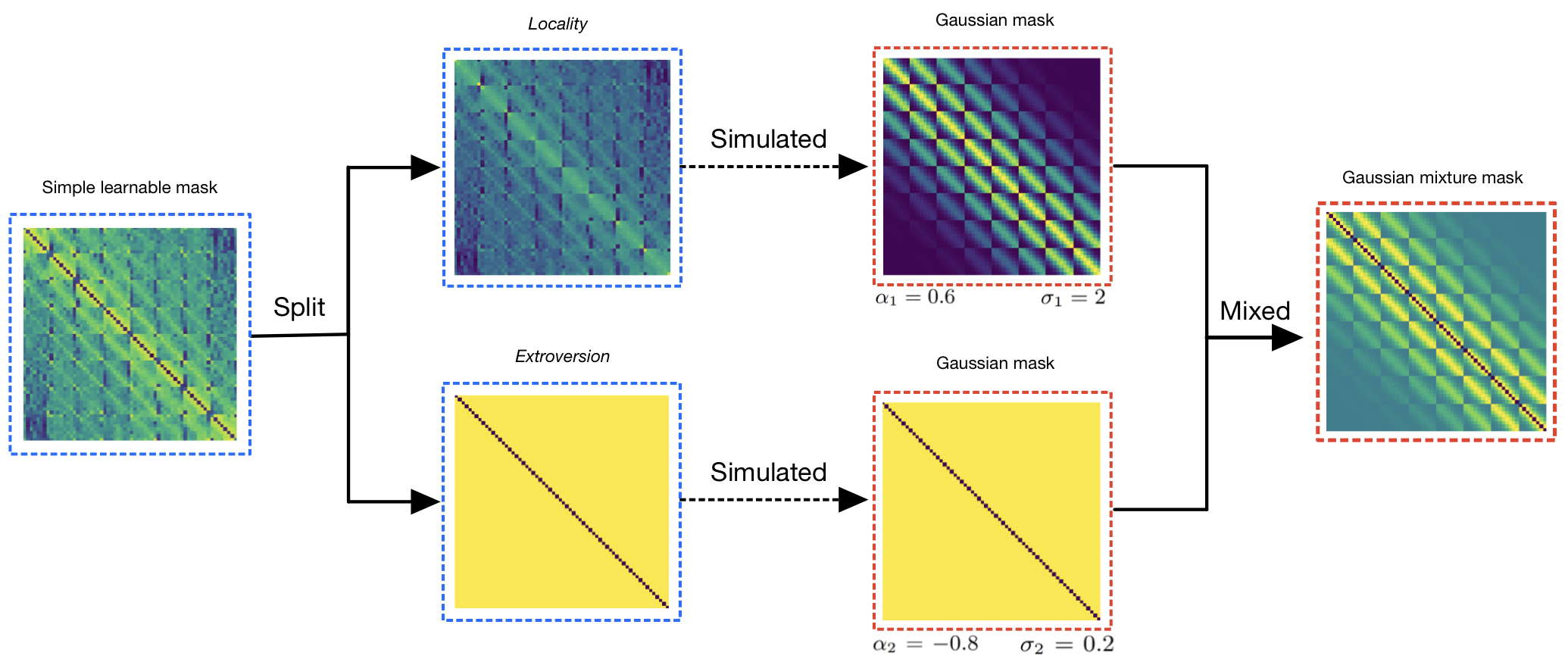}
    \caption{The \textbf{leftmost} figure shows the result of learning the simple learning mask, while the \textbf{rightmost} figure shows the result of the Gaussian mask mixture. The \textbf{middle} figure shows the fitting process, which first divides the simple learnable model into two parts, and then respectively fits them with two Gaussian masks, one with parameters $\alpha_2=-0.8$ and $\sigma_2=0.2$, and the other with parameters $\alpha_1=0.6$ and $\sigma_1=2$. The two features can be simultaneously approximated by two Gaussian masks.}
    \label{fig:simulation}
\end{figure*}

\subsection{Gaussian Mixture Mask}\label{sec:GMM}

We first define $K$ Gaussian weight matrices, each of which has two learnable parameters $\alpha_k$ and $\sigma_k$. The size of each Gaussian weight matrix is $N_{\text{patch}}^{\frac{1}{2}} \times 2 - 1$. $x$ and $y$ are the offsets of the horizontal and vertical coordinates from the center point, respectively. The size of each location in the Gaussian weight matrix is expressed as:

\begin{equation}
M_{x y}=\sum_{k=1}^K \alpha_k e^{-\frac{x^2+y^2}{2 \sigma_k^2+\epsilon}} \footnote{In order to prevent overflow, a very small value $\epsilon$ is added.} \;\;\;\; M \in \mathbb{R}^{N \times N}
\end{equation}

Let $x$ and $y$ be two variables, whose range lies between $(-N_{\text{patch}}^{\frac{1}{2}}, N_{\text{patch}}^{\frac{1}{2}})$. The GMM is created through the coalescence of $K$ Gaussian weight matrices. Then, a window of size $N_{\text{patch}}^{\frac{1}{2}}$ commences a sliding movement from the lower right corner to the upper right corner, and the resulting output is unfolded row by row. The unfolded results are subsequently spliced in the row direction.

The mapping between the horizontal coordinate $i$ and the vertical coordinate $j$ of the GMM after unfolding and the center offset of the horizontal coordinate $x$ and the vertical coordinate offset $y$ before unfolding can be formulated as:

\begin{equation}
\begin{aligned}
& x=\left|i \% N_{\text{patch}}^{\frac{1}{2}}-j \% N_{\text{patch}}^{\frac{1}{2}}\right| \\
& y=\left|i / / N_{\text{patch}}^{\frac{1}{2}}-j / / N_{\text{patch}}^{\frac{1}{2}}\right|
\end{aligned}
\end{equation}

In Figure~\ref{fig:simulation}, we illustrate how two Gaussian masks can be employed in a manual way to fit the learned ELM that was shown in Figure~\ref{fig:attention_map}. We first extract the extroversion trait from the ELM to obtain a mask, with the remaining contributing to the locality mask. Extroversion can be obtained with a small $\sigma$ and a negative $\alpha$, which constitutes a particular case of a Gaussian mask. Thereafter, we fit locality using a Gaussian mask, and finally, blend the two masks together to obtain a manually tailored GMM that incorporates both the traits of locality and extroversion.

\subsection{GMM Attention}\label{sec:GMMAtten}

We simply add the obtained GMM to the preceding softmax operation and verify it through experiments. Adding it before and after the softmax has no significant impact on the result. In order to maintain a structure similar to that of the original paper, we treat this operation as a simple masking operation to obtain the Gaussian mixture attention mechanism. In order to improve the generalizability of GMM, we employ Multi-Head GMM Attention, which consists of an independent dynamic GMM mask for each head in self-attention mechanism.

\begin{figure}[h] 
\centering 
\includegraphics[width=0.32\textwidth]{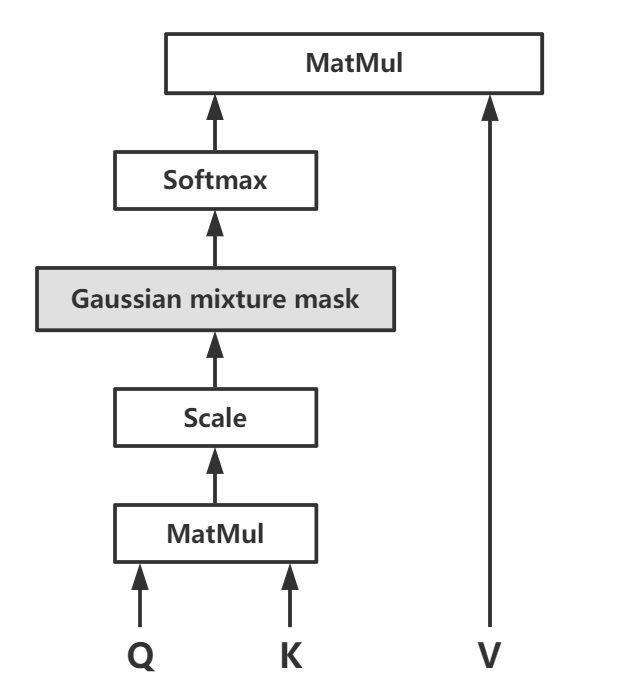} 
\caption{Gaussian Mixture Attention Mechanism} 
\label{fig:img4}
\end{figure}

We abandon the use of class tokens and adopt global pooling instead. This is mainly done to keep the attention weights matrix regular while reducing the computational parameters and not sacrificing the performance of the model. The feature vector $x_p \in \mathbb{R}^{N \times D}$ is used as input to the multi-head Gaussian mixture attention part, and is projected through three projection matrices $W_q, W_k, W_v \in \mathbb{R}^{D \times D}$ and corresponding biases $b_q, b_k, b_v \in \mathbb{R}^D$ to obtain three matrices $Q$ (query), $K$ (key) and $V$ (value).

\begin{align}
Q &= W_q x_p + b_q \\
K &= W_k x_p + b_k \\
V &= W_v x_p + b_v 
\end{align}

Then, the attention weights matrix can be computed as:

\begin{align}
A &= \frac{Q K^T}{\sqrt{d_k}} \quad \;\;\;\; A \in \mathbb{R}^{N \times N} 
\end{align}

with the position corresponding to the multiplication

\begin{align}
B_{i j} &= A_{i j} \circ M_{i j} 
\end{align}

Finally, after Sofamax and another linear projection, the result is obtained.

\begin{align}
\text{out} = W_o \operatorname{Softmax}(B)V + b_o
\end{align}

Overall:

\begin{equation}
\text{GMMAttention}(Q, K, V)\\
=\operatorname{Softmax}\left(\frac{QK^T}{\sqrt{d_k}} \circ M\right)V
\end{equation}

\textbf{Algorithm:}

The algorithm construction of GMM is as shown in Algorithm~\ref{alg:GMM}:

\begin{algorithm}[h]
\caption{Gaussian Mixture Mask}
\KwIn{Number of patches $N$, number of kernels $K$}

$\alpha_k = \mu_\alpha + \sigma_\alpha z$,  
$\sigma_k = \mu_\sigma + \sigma_\sigma z$\\
\hspace{12em}where $z \sim \mathcal{N}(0, 1)$\\

$M_{K \times N \times N} = \mathbf{0}$

\For{i = 0 to N-1}{
\For{j = 0 to N-1}{
$\Delta_x = i \% N^{\frac{1}{2}} - j \% N^{\frac{1}{2}}$ \\
$\Delta_y = i // N^{\frac{1}{2}} - j // N^{\frac{1}{2}}$ \\
\For{k = 0 to K-1}
{$M_{i j} += \alpha_k e^{-\frac{\Delta_x^2+\Delta_y^2}{2 \sigma_k^2+\epsilon}}$\\}
}
}
\Return{$M$}
\label{alg:GMM}
\end{algorithm}

\section{Experiments}

Our experiments mainly carried out on the small-scale datasets such as CIFAR-10, CIFAR-100, SVHN and Tiny-ImageNet. Firstly, we conducted pruning experiments on a wide range of ViTs in Section~\ref{sec:main_experiments}. Secondly, we conducted experiments on the settings of GMM hyperparameters in Section~\ref{sec:hypParameter}. Lastly, we compared the applicability performance of GMM in different visual deformations in Section~\ref{sec:Variants}, focusing on the comparison between GMM and the mainstream local hierarchical ViT-Swin and deep ViT-CaiT.

\subsection{Main Results}\label{sec:main_experiments}

\begin{table*}[t]
    \centering
    \scalebox{1.0}{
    \begin{tabular}{l|cccc|ccc}
    \toprule
    \textbf{Model}              & \textbf{CIFAR-10} & \textbf{CIFAR-100} & \textbf{SVHN} & \textbf{Tiny-ImageNet}& \textbf{Parameters} & \textbf{MACs} & \textbf{Depth}\\ 
    \midrule
    ViT             & 93.65\% & 75.36\% & 97.93\% & 59.89\% & 2.7M & 170.9M & 9\\
    GMM-ViT         & \textbf{95.06\%} & \textbf{77.81\%} & \textbf{98.01\%} & \textbf{62.27\%} & 2.7M& 170.9M  & 9\\
    \midrule
    Swin            & 95.26\% & 77.88\%  & 97.89\% & 60.45\%  & 7.1M & 236.9M & 12\\
    GMM-Swin        & \textbf{95.39\%} & \textbf{78.26\%} & \textbf{97.90\%} & \textbf{61.03\%} & 7.1M & 236.9M & 12 \\
    \midrule
    CaiT            & 94.79\% & 78.42\% & 98.13\% & 62.46\% & 5.1M & 305.9M & 26\\
    GMM-CaiT        & \textbf{95.15\%} & \textbf{78.97\%} & 98.09\% & \textbf{63.64\%} & 5.1M & 305.9M & 26\\ 
    \midrule
    PiT            & 93.68\% & 72.82\% & 97.78\% & 57.63\% & 7.0M & 239.1M & 12\\
    GMM-PiT        & \textbf{94.41\%} & \textbf{74.16\%} & \textbf{97.82\%} & \textbf{58.37\%} & 7.0M & 239.1M & 12\\ 
    \midrule
    T2T            & 95.32\% & 78.10\% & 97.99\% & 61.50\% & 6.5M & 417.4M & 13\\
    GMM-T2T        & \textbf{96.16\%} & \textbf{79.91\%} & 97.98\% & \textbf{63.33\%} & 6.5M & 417.4M & 13\\ 
    \bottomrule
    \end{tabular}
    }
    \vspace{0.8em}
    \caption{Top-1 accuracies of different ViTs and GMM variants obtained on small datasets (\%).}
    \label{tab:overall}
\end{table*}

\noindent\textbf{Experimental setup}. We implemented image classification Top-$1$ accuracy experiments on the small datasets CIFAR-10, CIFAR-100, SVHN, and Tiny-ImageNet using the Timm library\cite{rw2019timm} and Vision Transformer for Small-Size Datasets \cite{lee2021vision}. The base configuration and training configuration of the vision transformer were adopted for all experiments. Specifically, the patch size was determined based on the size of the input image: $4$ for $32 \times 32$ images (CIFAR-10, CIFAR-100, and SVHN) and $8$ for $64 \times 64$ image (Tiny-ImageNet). For training regime,  CutMix \cite{yun2019cutmix}, Mixup \cite{zhang2017mixup}, Auto Augment \cite{cubuk2019autoaugment}, Repeated Augment \cite{cubuk2020randaugment}, Label Smoothing \cite{szegedy2016rethinking}, Stochastic Depth \cite{huang2016deep}, Random Erase \cite{zhong2020random}, AdamW \cite{kingma2014adam}, and Cosine Learning Rate Scheduler \cite{loshchilov2016sgdr} were all employed.\\

\noindent\textbf{Implement details.} GMM can theoretically be added to any ViTs with self-attention mechanism, but the implementation is different among different variants. In the vanilla ViT~\cite{dosovitskiy2020image}, to conform the GMM matrix, the class token is dropped and the global pool is used as the final classifier. In Swin~\cite{liu2021swin}, GMM is applied to each attention window, and the size of the GMM matrix is determined by the size of the window. In the Shifted window attention module of Swin, GMM is placed after the scaling operation and before applying the shifted window attention mask. In CaiT~\cite{touvron2021going}, the specialty is the last class attention layer and GMM is only adopted in the self-attention layers preceding the class attention layer. In PiT~\cite{heo2021rethinking}, the class token is retained and the size of the GMM matrix varies from level to level. In the implementation of T2T~\cite{yuan2021tokens}, the T2T Transformer structure is preserved, and GMM is inserted directly in the self-attention. On the last transformer structure, the class token is dropped and replaced with the global pooling, and GMM is added. The results of above five VT and its GMM variants are presented in the table~\ref{tab:overall}.

\subsection{Hyperparameter Setting}\label{sec:hypParameter}

\noindent\textbf{Parameter initialization.} Through simple experiments without deliberate adjustment, it was found that when the learnable parameters $\alpha$ and $\sigma$ are initialized, $\alpha$ should be initialized to a normal distribution with mean 0 and standard deviation 2, and $\sigma$ should be initialized to a normal distribution with mean 10 and standard deviation 10, which yields better results.

$$
    \alpha \sim \mathcal{N}(0, 4)
$$
$$
    \sigma \sim \mathcal{N}(10, 100)
$$

\noindent\textbf{The number of Gaussian mixture kernels} applied in the model will affect its local modeling ability and effectiveness. We train GMM-ViT on the CIFAR-100 and Tiny-ImageNet datasets, controlling the number of masks for each GMM from 1 to 10. Experiments on small datasets show that as the number of Gaussian kernels in the mixture model starts to increase from one, the precision of the model will quickly rise and eventually reach saturation. When an appropriate amount of mask is applied to each layer, the generalization performance of the model will be better on different types of dataset.

\begin{figure}[h]
    \centering
    \includegraphics[width=\linewidth]{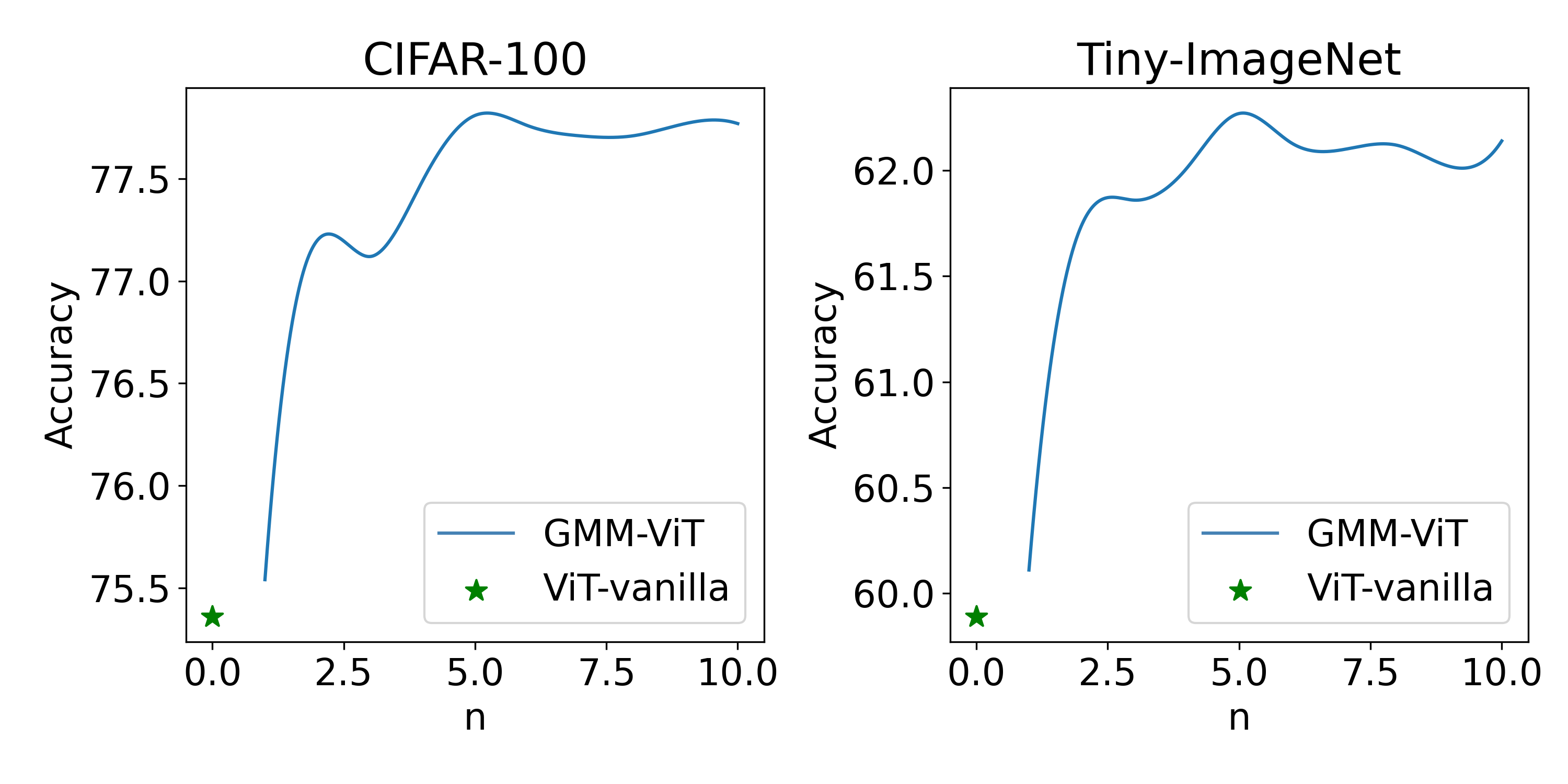}
    \caption{Comparison of Top-1 Accuracy of GMM-ViT with different numbers of Gaussian kernels on CIFAR100 and Tiny-ImageNet datasets. $n$ is the number of GMMs, represents the standard ViT model with a GMM composed of $n$ layers of Gaussian kernels.}
    \label{fig:line_plots}
\end{figure}

The line plots in Figure \ref{fig:line_plots} show that insufficient Gaussian kernels cannot fulfil the local modelling requirements of the model. When the number of Gaussian kernels reaches around 5, the Gaussian kernels reach a redundant state in terms of performance on small dataset, and further increasing the number of Gaussian kernels will not improve the performance of the model.

\subsection{GMM for Different Variants of ViT}\label{sec:Variants}

\noindent\textbf{Comparison to Swin.} After adding a GMM with 150 parameters to a 2.5M ViT, the performance on CIFAR10 is close to a 7.1M Swin Transformer. Applying the GMM to the Swin model can still increase the accuracy of the Swin model.

\setlength{\tabcolsep}{3mm}
\begin{table}[h]
  \centering
    \scalebox{0.73}{
    \begin{tabular}{c|c|c}
    \toprule
    \textbf{Model} & \textbf{Top-1 Acc} & \textbf{Number of Parameters} \\
    \midrule
    ViT & 93.65\% & 2.5M \\
    Swin & 95.11\% & 7.1M \\
    \midrule
    GMM-ViT & 95.06\% & 2.5M (+150 params) \\
    GMM-Swin & 95.42\% & 7.1M (+240 params) \\
    \bottomrule 
    \end{tabular}% 
    }
  \vspace{0.8em}  
  \caption{Performance comparison of four models on CIFAR-10. Swin performs better than ViT but at a cost of increasing the parameters by more than 3 times. Lightweight GMM module help ViT achieves almost the same performance as Swin by increasing only 150 parameters. Swin can further be imrpoved by our GMM module.}
  \label{tab3}%
\end{table}%

From the table \ref{tab3}, we can see that the Top-1 accuracy of the GMM-ViT model is 95.06\%, which is higher than the standard ViT model of 93.65\%, and the GMM-Swin model's Top-1 accuracy is 95.42\%, which is higher than the standard Swin model's 95.11\% accuracy. This proves that the GMM model has a strong effect in improving the accuracy of ViT and Swin models. In addition, the number of parameters of GMM-ViT and GMM-Swin models only increased by 150 and 240, respectively, which shows the advantage of the GMM model with low parameter quantity.
\\\\
\noindent\textbf{GMM in deep ViTs.} Table \ref{tab:vit_deep} investigates the impact of GMM on ViTs with various depths, under a comparable parameter scale. Increasing the number of layers and reducing the dimension of feature vectors, the accuracy of deep ViT models drops significantly; GMM can significantly improve this situation, with the accuracy of the 60-layer ViT pulled back to the level of the 15-layer standard ViT after adding the GMM. 

\begin{table}[h]
\centering
\scalebox{0.73}{
\begin{tabular}{ccccc}
\toprule
\textbf{Model} & \textbf{CIFAR-10} & \textbf{Depth} & \textbf{Hidden-dim} & \textbf{\#Params} \\
\midrule
ViT-base & 94.32\% & 6 & 252 & 3,091,798 \\
GMM-ViT & \textbf{94.68\%} & 6 & 252 & 3,091,858 \\
\midrule
ViT-base & 94.17\% & 9 & 192 & 2,692,042 \\
GMM-ViT & \textbf{94.87\%} & 9 & 192 & 2,692,096 \\
\midrule
ViT-base & 93.65\% & 15 & 144 & 2,523,610 \\
GMM-ViT & \textbf{95.06\%} & 15 & 144 & 2,523,760 \\
\midrule
ViT-base & 93.60\% & 30 & 108 & 2,838,790 \\
GMM-ViT & \textbf{94.54\%} & 30 & 108 & 2,838,970 \\
\midrule
ViT-base & 90.91\% & 60 & 72 & 2,531,890 \\
GMM-ViT & \textbf{93.32\%} & 60 & 72 & 2,532,250 \\
\bottomrule
\end{tabular}
}

\vspace{0.8em}
\caption{Performance comparison of ViT with different depths on CIFAR-10 by controlling the dimension of the hidden vector to maintain the order of magnitude of parameters changed due to the change of layers.}
\label{tab:vit_deep}
\end{table}

\begin{figure*}[b]
    \centering
    \includegraphics[width=\linewidth]{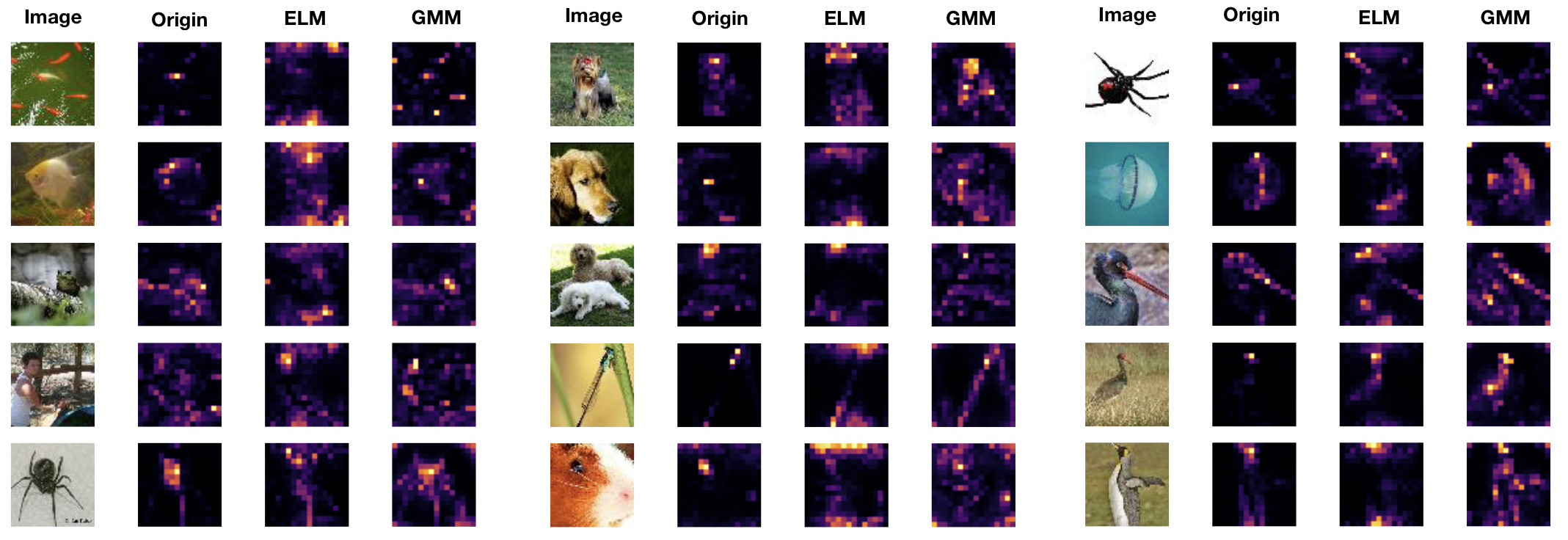}
    \caption{
In each image group, the \textbf{left} image shows the input image, the \textbf{second left} image shows the standard ViT's attention map, the \textbf{second right} image represents the ViT's attention map with ELM strengthening, and the \textbf{rightmost} image shows the ViT's attention map with GMM strengthening.}
    \label{fig:enter-label}
\end{figure*}

GMM has the plasticity to increase the model depth as well. A GMM can be separately generated to contain a small $\sigma$ and a dynamic changed $\alpha$ to modulate the weight of its patch information. This kind of mask can in some way enhance the residual connection or offset some negative effects brought by the residual connection. Self-attention mechanism tends to act locally in shallow networks and globally in deep networks~\cite{dosovitskiy2020image,li2022locality}. In deep networks, the residual connection helps to make the next layer contain more information from the previous layer, thus stabilizing the learning process. This kind of high-sigma mask can make the patch information contain more of its own information after updating, thus achieving a similar effect. In shallow networks, this kind of self-tending network may delay the learning process. At this time, a high-sigma negative-alpha mask can offset this effect, thus dynamically regulating the information interaction between layers.

\begin{table}[h]
\centering
\scalebox{0.73}{
\begin{tabular}{ccccc}
\toprule
\textbf{Model} & \textbf{CIFAR-10} & \textbf{Depth} & \textbf{Hidden-dim} & \textbf{\#Params} \\
\midrule
CaiT & 95.38\% & 26(2SA+24CA) & 256 & 9,020,618 \\
GMM-ViT & \textbf{95.45\%} & 30 & 192 & 8,917,750 \\
\bottomrule
\end{tabular}
}
\vspace{0.8em}
\caption{Comparison of performances of GMM-ViT and CaiT at the same level of parameters and depth on CIFAR-10.}
\label{tab:Gmm_CaiT}
\end{table}

In comparison with CaiT, GMM-ViT as shown in Table~\ref{tab:Gmm_CaiT} can converge to a higher accuracy at deeper layers under the same parameter and depth level.

\section{ELM \textit{v.s.} GMM}\label{sec:4.3}

\noindent\textbf{Attention map in three ways.} We trained three versions of ViT on Tiny-ImageNet: standard ViT, ViT with ELM, and ViT with GMM, and set the patch size to 4. The top-1 accuracy is shown in Table~\ref{tab:Attention_map}.

\begin{table}[h]
\centering
\scalebox{0.7}{
\begin{tabular}{l|c|r}
\toprule
\textbf{Model} & \textbf{Top-1 Acc on Tiny-Imagenet} & \textbf{Params}\\ 

\midrule
ViT & 61.67\% (+0.00\%) & 2,469,128 (+0)  \\
ELM-ViT & 63.26\% (+1.59\%) & 2,993,416 (+524,288)  \\
GMM-ViT & \textbf{64.41\% (+2.74\%)} & 2,469,608 \textbf{(+480)}  \\

\bottomrule
\end{tabular}
}
\vspace{1em}
\caption{Comparison of performance between GMM and ELM on Tiny-ImageNet}
\label{tab:Attention_map}
\end{table}

We visualized the last-layer attention maps of three versions of ViT on ImageNet dataset shown in Figure~\ref{fig:enter-label}, and concluded that GMM-ViT has stronger expressive power than ELM-ViT and Standard-ViT in terms of attention maps.

\noindent\textbf{Comparison of performance between GMM and ELM.} An ELM consists of $N \times N$ learnable parameters, which will undoubtedly show a quadratic growth as the patch size decreases or the image size increases. However, the learnable parameters of a single-layer GMM are $2\times K$, where $K$ is the number of Gaussian mixture kernels, which shows a linear growth with the increase of mask number. Therefore, the parameter scale of GMM is far less than that of a single-layer ELM.

\begin{table}[h]
\centering
\scalebox{0.8}{
\begin{tabular}{l|c|r}
\toprule
\textbf{Model} & \textbf{Top-1 Acc} & \textbf{\#Params}\\ 
\midrule
\textit{CIFAR-10 dataset}   \\
\midrule
ViT$_{d9}$ & 94.17\% (+0.00\%) & 2,692,042 (+0)  \\
ELM-ViT$_{d9}$ & 94.65\% (+0.48\%) & 2,728,906 (+36,864)  \\
GMM-ViT$_{d9}$ & \textbf{94.94\% (+0.77\%)} & 2,692,186 \textbf{(+144)}  \\
\cmidrule(l){1-3}
ViT$_{d15}$ & 93.65\% (+0.00\%) & 2,523,610 (+0)  \\
ELM-ViT$_{d15}$ & 94.65\% (+1.00\%) & 2,585,050 (+61,440)  \\
GMM-ViT$_{d15}$ & \textbf{95.06\% (+1.41\%)} & 2,523,760 \textbf{(+150)}  \\ 
\midrule
\textit{CIFAR-100 dataset}  \\
\midrule
ViT$_{d9}$ & 75.36\% (+0.00\%) & 2,709,412 (+0)  \\
ELM-ViT$_{d9}$ & 76.97\% (+1.61\%) & 2,746,276 (+36,864)  \\
GMM-ViT$_{d9}$ & \textbf{77.56\% (+2.20\%)} & 2,709,556 \textbf{(+144)}  \\
\cmidrule{1-3}
ViT$_{d15}$ & 74.20\% (+0.00\%) & 2,536,660 (+0)  \\
ELM-ViT$_{d15}$ & 76.37\% (+2.17\%) & 2,598,100 (+61,440)  \\
GMM-ViT$_{d15}$ & \textbf{77.61\% (+3.41\%)} & 2,536,810 \textbf{(+150)}  \\ 
\bottomrule
\end{tabular}
}
\vspace{1em}
\caption{Comparison of performance between GMM and ELM. Compared to ELM, GMM has fewer parameters but higher performance.}
\label{tab:perf_gmm_elm}
\end{table}

Table~\ref{tab:perf_gmm_elm} shows the performance comparison of two different versions of ViT, the first with 9 layers and 192 hidden dimensions and the second with 15 layers and 144 hidden dimensions, respectively applied with GMM and ELM. It can be seen that GMM still obtained better modeling ability than ELM with extremely few parameters added compared to ELM, which shows the feasibility of the GMM structure.\\

\section{Conclusions}

In this work, we revisit RetNet from the perspective of convolution and propose a method to enhance ViT training from scratch on small datasets. We initially introduce an Element-wise Learnable Mask to dynamically regulate the locality of the attention mechanism, which enhances ViTs training on small datasets from scratch and summarizes two characteristics of the mask. Based on these two characteristics, we further propose a Gaussian Mixture Mask, which has a much higher performance than the Element-wise Learnable Mask and reduces the cost of parameter and computation overhead to almost zero.

{\small
\bibliographystyle{ieee_fullname}
\bibliography{egbib}
}

\end{document}